\documentclass[letterpaper]{article} 
\usepackage[]{aaai25}  

\usepackage[dvipsnames]{xcolor}
\usepackage{times}  
\usepackage{helvet}  
\usepackage{courier}  
\usepackage[hyphens]{url}  
\usepackage{graphicx} 
\usepackage{natbib}  
\usepackage{caption} 
\usepackage{algorithm}
\usepackage{algorithmic}
\usepackage{longtable}

\usepackage{newfloat}
\usepackage{listings}
\DeclareCaptionStyle{ruled}{labelfont=normalfont,labelsep=colon,strut=off} 
\floatstyle{ruled}
\newfloat{listing}{tb}{lst}{}
\floatname{listing}{Listing}
\usepackage{natbib}
\usepackage{todonotes}
\usepackage{booktabs}
\usepackage{mathtools}
\usepackage{pifont}
\newcommand{\cmark}{\ding{51}}%
\newcommand{\xmark}{\ding{55}}%
\usepackage{amsmath, amssymb, amsthm, mathtools}

\usepackage{caption}
\usepackage{subcaption}
\usepackage{adjustbox}
\usepackage{graphicx}

\usepackage{physics}
\usepackage{tikz}
\usepackage{capt-of}
\usepackage{svg}
\usepackage[hang,flushmargin]{footmisc} 
\usepackage{hyperref}

\theoremstyle{remark}
\newtheorem*{remark*}{Remark}

\title{Discovering Symmetries of ODEs by Symbolic Regression}
\author {
    Paul Kahlmeyer,
    Niklas Merk,
    Joachim Giesen
}
\affiliations {
   Friedrich Schiller University Jena\\
    paul.kahlmeyer@uni-jena.de, niklas.merk@uni-jena.de, joachim.giesen@uni-jena.de
}

\begin{document}

\maketitle

\begin{abstract}
Solving systems of ordinary differential equations (ODEs) is essential when it comes to understanding the behavior of dynamical systems. Yet, automated solving remains challenging, in particular for nonlinear systems.  Computer algebra systems (CASs) provide support for solving ODEs by first simplifying them, in particular through the use of Lie point symmetries. Finding these symmetries is, however, itself a difficult problem for CASs. Recent works in symbolic regression have shown promising results for recovering symbolic expressions from data. Here, we adapt search-based symbolic regression to the task of finding generators of Lie point symmetries. 
With this approach, we can find symmetries of ODEs that existing CASs cannot find\footnote{Code: \href{https://github.com/kahlmeyer94/ODESym}{\textcolor{blue}{\texttt{https://github.com/kahlmeyer94/ODESym}}}}.
\end{abstract}

\section{Introduction}

The study of dynamical systems plays a key role in all natural sciences~\citep{strogatz_problems}, as well as in economics~\citep{lorenz1993_economics} and in the social sciences~\citep{Richardson2014_psychology}. Describing the state of dynamical systems explicitly as a function over time, that is, in integral form, is usually difficult. It is often much easier to describe the system by the change of its state over time, that is, in differential form. Systems of ordinary differential equations (ODEs) are such a description in differential form. Systems of ODEs are a key mathematical concept for modeling and understanding dynamical systems \citep{Smale1974_ODE, tenenbaum1985_ODE}. Transforming the differential form into integral form means solving an ODE system, that is, finding an explicit function that satisfies the ODE. Solving ODEs is crucial for predicting system behavior and for analyzing its dynamics. A large part of research has been devoted to finding solutions to ODEs by numerical or analytical methods. Numerical solutions involve simulating the system from a specific initial condition, offering effective solutions for a wide range of ODEs, but lacking the interpretability of analytical solutions.

\begin{figure}
    \centering
    \includegraphics[width = \columnwidth]{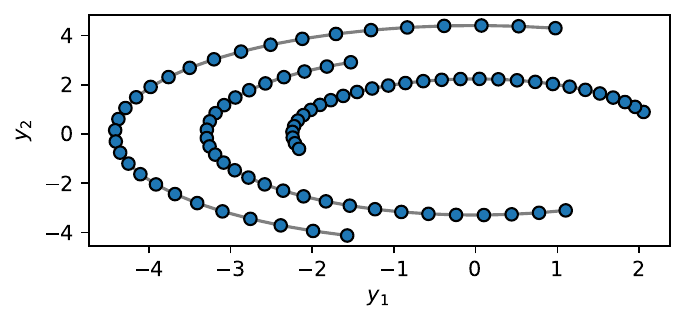}
    \caption{Three numerical solutions for the example ODEs $y' = [-y_2, y_1]$ in state space. Time steps are equally spaced between two adjacent points of a single trajectory.}
    \label{fig:example_intro}
\end{figure}

For an example, consider a system, where we observe a quantity $y = [y_1, y_2]$ over time $t$. The dynamics of the system is such that the change of the first component $y_1$ is negatively correlated with the magnitude of the second component $y_2$, whereas the change of the second component is positively correlated with the magnitude of the first component. This simple system can be modeled by the following system of two linear ODEs
\begin{align*}
    y_1' &= -y_2\\
    y_2' &= y_1\,
\end{align*}
where $y_i'$ is the derivative of the time-dependent function $y_i$. Three numerical solutions for the example system are shown in Figure~\ref{fig:example_intro}. Note that these solutions consist only of sample points.
The system is simple enough that we can also provide an analytical solution
\begin{align*}
    y_1(t) &= a\cdot \cos(t + b)\\
    y_2(t) &= a\cdot \sin(t + b)
\end{align*}
for some constants $a$ and $b$. The three numerical solutions in Figure~\ref{fig:example_intro} correspond to different choices of the constants $a$ and $b$.

Many natural phenomena, however, are described by nonlinear systems of ODEs, for instance chaotic systems. Nonlinear ODEs are typically much more difficult to solve analytically. Solutions are known only in special cases, when the nonlinearities can be resolved by transforming the system into an equivalent system of linear ODEs, such as Riccati or Bernoulli differential equations. Sophus Lie~\citep{Lie1888} proposed a fundamental and unifying framework for the analysis of systems of ODEs using symmetries. A symmetry of an ODE system describes a mapping from one solution to another one. Symmetries can be used to derive a coordinate transformation. In the transformed coordinate system, the number of equations of a system of ODEs is reduced by one. Sometimes the reduced system is easier to solve analytically, or it can be reduced further by exploiting another symmetry. Our linear example system has a particularly simple symmetry, it is invariant under positive scaling. Therefore, a parameterized family of symmetries is given by
\[
h_\theta(t, y_1, y_2) = \left(t, e^{\theta}y_1, e^{\theta}y_2\right) =: \left( \hat t, \hat y_1, \hat y_2 \right),
\]
where $\theta \in\mathbb{R}$ parameterizes the family. This is easily verified, because
\[
e^{\theta}y_1(t) = \hat a\cdot\cos(t+b)
\quad\textrm{and}\quad
e^{\theta}y_2(t) = \hat a\cdot\sin(t+b),
\]
with $\hat a = e^{\theta} a$, is a solution of the ODE 
\begin{align*}
    \hat y'_1 &= -\hat y_2 \\ 
    \hat y'_2 &= \hat y_1, 
\end{align*}
where $\hat y'_i$ is the derivative of $\hat y_i$ with respect to $\hat t$. In the supplement, we demonstrate how this family of symmetries can be used to simplify the example system. For nonlinear systems, the challenge becomes to find the symmetries.

Given an ODE system in symbolic form, the necessary condition for a symmetry can be expressed as a system of partial differential equations (PDEs). Traditionally, CASs have been used to solve the PDE system symbolically. This approach however is again limited by the capabilities of CASs to tackle nonlinear systems. In this work, we propose to use symbolic regression to search directly for expressions that satisfy the symmetry condition. Unlike traditional regression methods, symbolic regression does not fixate on a given class of functions, but rather searches in the space of \textit{all} possible functions that can be constructed from a given vocabulary of operators. Recent progress in symbolic regression has produced efficient search algorithms that in contrast to previous work in symbolic regression can also be used to search for symmetries of ODE systems. We demonstrate that symbolic regression is able to find symmetries of nonlinear systems where the traditional CAS approach fails, making it a valuable extension for any ODE related toolbox. 

\section{Related Work}

We organize our discussion of related work along three lines of research, namely, computer algebra systems for simplifying and solving ODEs, symbolic regression, and the related field of learning conserved quantities and invariants of dynamical systems.

\subsection {Computer Algebra Systems}

Computer algebra systems (CASs) are software programs designed for performing symbolic mathematical operations. Unlike numerical software, which primarily deals with numerical values and calculations, CASs manipulate mathematical expressions and symbols directly, enabling tasks such as algebraic simplification, equation solving, symbolic differentiation, and symbolic integration. The ability of CASs to solve ODEs or the corresponding symmetry PDEs relies on specialized algorithms for detecting systems with specific structures, such as, linear, Riccati-type, or Bernoulli-type, which have known solutions. 

Examples of CASs include the commercial software systems Mathematica~\citep{mathematica} and Maple~\citep{maple}. Both systems provide a wide range of functionalities for analyzing ODEs and their symmetries.
Maple supports symbolic symmetry finding for ODEs through the symgen package~\citep{Chebterrab1995_symgen,symgen97,symgen98}, which iteratively tests for special structures, but notably cannot handle systems of ODEs. 
For Mathematica, two packages for finding symmetries exist: (1) the SYM package~\citep{Stylianos2005_SYM} and (2) the MathLie package~\citep{Baumann2002_MathLie}. Both packages can handle systems of ODEs and compute symmetries by deriving the corresponding partial differential equations and trying to solve them.

\subsection{Symbolic Regression}

Given a training set of labeled data, the goal in symbolic regression is to find an interpretable expression, where the underlying function generalizes well beyond the training data.
Symbolic regression algorithms differ in the way they search the space of expressions. 
The majority of algorithms follow the \emph{genetic programming} paradigm~\citep{koza1994genetic,Holland:1975}. The paradigm has been implemented in the seminal Eureqa system by~\citet{eureqa_lipson09} and in gplearn by \citet{gplearn_stephens16}. More recent implementations are provided by~\citet{eplex_lacava16,kommenda2020parameter,virgolin2021improving}. The basic idea is to create a population of expression trees and to recombine the best performing trees into a new population by exchanging subtrees. 
Alternatively, Bayesian inference methods leverage Markov Chain Monte Carlo (MCMC) sampling to draw expressions from a posterior distribution, with variations in the choice of prior distribution.
\citet{jin2020bayesian} use a hand-designed prior distribution on expression trees, whereas \citet{bayesianscientist_guimera20} compile a prior distribution from a corpus of mathematical expressions.
A different branch of work focuses on using deep learning:
\citet{dsr_petersen21} and \citet{dsr2_petersen21} train recurrent neural networks, \citet{transformer_kamienny22} train a transformer for generating expression trees token-by-token in preorder. 
Recently, \citet{dsr3_petersen22} have combined different symbolic regressors based on genetic programming, reinforcement learning, and transformers together with problem simplification into a unified symbolic regressor. 

Overall, the idea of an unbiased, more or less assumption-free search has shown the greatest potential when it comes~to recovering expressions from data. The AIFeynman project~\citep{feynmanAI_udrescu20} uses a set of neural-network-based statistical property tests to scale an unbiased search to larger search spaces. Statistically significant properties, for example, additive or multiplicative separability, are used to decompose the search space into smaller parts, which can be addressed by brute-force search.
Recently, \citet{kahlmeyer2024udfs} have proposed an unbiased search for expressions represented by directed, acyclic graphs (DAGs).
Unlike the more popular expression tree representation, DAGs allow for criterion-based search considering all output dimensions simultaneously.

\subsection{Conserved Quantities and Invariants}

Conserved quantities and invariants are both closely related to symmetries. 
In a celebrated theorem, \citet{Noether1918} has shown a correspondence between continuous symmetries and conservation laws for physical systems. 
\citet{liu21} have designed and implemented the AI Poincar\'{e} algorithm for directly learning the number and, by using their symbolic regressor AIFeynman~\citep{feynmanAI_udrescu20}, the symbolic form of conserved quantities from time series data. 
The extension Poincar\'{e}~2.0 by \citet{LiuT22aipoincare} uses numerically generated time series data for learning $n$-th order conserved quantities from ODEs and some PDEs. 
\citet{liu22} generalize this work and propose an automated workflow for the discovery of a set of six predefined symmetry classes by learning a coordinate transformation that makes these symmetries apparent. The coordinate transformation is approximated by a neural network and translated into a symbolic form using AIFeynman.
\citet{Bondesan19} follow a similar approach and learn a coordinate transformation, which makes it easier to identify conserved quantities. 

Our proposed approach differs from the aforementioned work in two major aspects. 
First, conserved quantities directly correspond to a special kind of symmetry, namely, the Hamiltonian symmetry. While it can reveal interesting properties of the underlying system, we explicitly exclude these symmetries in this work, as they cannot be used to derive the coordinate transformation that simplifies the system.
Second, we do not use an intermediary approximation by some black box module like a neural network, but apply symbolic regression directly to the underlying system.

\section{Symmetries of ODEs}

Before we can describe our approach for learning symmetries for ODEs, we briefly review ODEs and related symmetry concepts that are used in our approach. For a more detailed treatment of the mathematical background, we recommend the books by \citet{bluman89}, \citet{stephani93}, or~\citet{Lie1888}.

A first order ODE system is a system of equations of the form
\[
y' \,=\, f(t,y),
\]
where $y'$ denotes the derivative of the function 
\[ 
y:U\subseteq \mathbb{R}\rightarrow\mathbb{R}^d,\, t\mapsto y(t)
\]
and $f$ is a function $f:V\subseteq\mathbb{R}^{d+1}\rightarrow\mathbb{R}^d$. Considering only first order ODEs is not really a restriction. Any higher order ODE system can be expressed by an equivalent first order ODE system using additional variables. 

A (point) symmetry transformation of an ODE system is a differentiable, invertible map
\[
h:(t,y) \mapsto \left(\hat{t}(t,y),\hat{y}(t,y)\right),
\]
which maps each solution $y(t)$ of the ODE to a solution $\hat{y}(\hat{t})$. For a differentiable symmetry transformation $h$, the symmetry condition, which requires that also $\hat{y}(\hat{t})$ is a solution of the ODE, is not difficult to check. However, the search space for symmetry transformations is vast, yet not unstructured. The set of symmetry transformations for an ODE has a group structure, where the group operation is the composition of transformations and the neutral element is the identity transformation. For solving ODEs, one-parameter subgroups of the symmetry group, namely one-parameter Lie groups, are particularly interesting. In our approach, the function $f$ is known. Therefore, we can sample data from $f$ and then use the data to learn Lie symmetries of the corresponding ODE system.     

A one-parameter Lie group is a family of symmetry transformations
\[
h_\theta:(t,y) \mapsto \big(\hat{t} (\theta, t, y),\hat{y}(\theta, t, y) \big)
\]
that are smooth in $t$ and $y$, and analytical in the parameter $\theta \in \mathbb{R}$. The group structure needs to satisfy 
\[
h_\theta \circ h_{\theta'} = h_{\theta +\theta'} \quad\textrm{and}\quad h_1 = \mathrm{id}.
\]

\begin{figure}[h]
\centering
\tikzset{every picture/.style={line width=0.75pt}} 
\begin{tikzpicture}[x=0.75pt,y=0.75pt,yscale=-0.5,xscale=0.5]
\draw  [dash pattern={on 0.75pt off 1.5pt}]  (141.17,293.83) .. controls (170.17,265.83) and (256,117.5) .. (263,65.5) ;
\draw  [dash pattern={on 0.75pt off 1.5pt}]  (226.33,293.5) .. controls (265.33,250.5) and (357,113.5) .. (359,67.5) ;
\draw  [dash pattern={on 0.75pt off 1.5pt}]  (350.67,294.17) .. controls (387,252.5) and (461,126.5) .. (467,65.5) ;
\draw    (361.67,280.42) -- (374.78,263.5) ;
\draw [shift={(376,261.92)}, rotate = 127.77] [fill={rgb, 255:red, 0; green, 0; blue, 0 }  ][line width=0.08]  [draw opacity=0] (12,-3) -- (0,0) -- (12,3) -- cycle    ;
\draw    (380.83,253.25) -- (393.06,234.75) ;
\draw [shift={(394.17,233.08)}, rotate = 123.47] [fill={rgb, 255:red, 0; green, 0; blue, 0 }  ][line width=0.08]  [draw opacity=0] (12,-3) -- (0,0) -- (12,3) -- cycle    ;
\draw    (448.15,129.13) -- (456.1,109.11) ;
\draw [shift={(456.83,107.25)}, rotate = 111.65] [fill={rgb, 255:red, 0; green, 0; blue, 0 }  ][line width=0.08]  [draw opacity=0] (12,-3) -- (0,0) -- (12,3) -- cycle    ;
\draw    (419.17,188.92) -- (429.22,170.18) ;
\draw [shift={(430.17,168.42)}, rotate = 118.22] [fill={rgb, 255:red, 0; green, 0; blue, 0 }  ][line width=0.08]  [draw opacity=0] (12,-3) -- (0,0) -- (12,3) -- cycle    ;
\draw    (399.99,222.79) -- (411.97,202.56) ;
\draw [shift={(412.99,200.84)}, rotate = 120.64] [fill={rgb, 255:red, 0; green, 0; blue, 0 }  ][line width=0.08]  [draw opacity=0] (12,-3) -- (0,0) -- (12,3) -- cycle    ;
\draw    (459.67,98.42) -- (465.29,78.67) ;
\draw [shift={(465.83,76.75)}, rotate = 105.89] [fill={rgb, 255:red, 0; green, 0; blue, 0 }  ][line width=0.08]  [draw opacity=0] (12,-3) -- (0,0) -- (12,3) -- cycle    ;
\draw    (435.67,156.92) -- (444.66,137.4) ;
\draw [shift={(445.5,135.58)}, rotate = 114.75] [fill={rgb, 255:red, 0; green, 0; blue, 0 }  ][line width=0.08]  [draw opacity=0] (12,-3) -- (0,0) -- (12,3) -- cycle    ;
\draw    (238.75,133.25) -- (248.62,111.38) ;
\draw [shift={(249.44,109.56)}, rotate = 114.29] [fill={rgb, 255:red, 0; green, 0; blue, 0 }  ][line width=0.08]  [draw opacity=0] (12,-3) -- (0,0) -- (12,3) -- cycle    ;
\draw    (250.91,105.09) -- (260.13,83.3) ;
\draw [shift={(260.91,81.45)}, rotate = 112.93] [fill={rgb, 255:red, 0; green, 0; blue, 0 }  ][line width=0.08]  [draw opacity=0] (12,-3) -- (0,0) -- (12,3) -- cycle    ;
\draw    (191.15,222.05) -- (203.09,201.94) ;
\draw [shift={(204.11,200.22)}, rotate = 120.7] [fill={rgb, 255:red, 0; green, 0; blue, 0 }  ][line width=0.08]  [draw opacity=0] (12,-3) -- (0,0) -- (12,3) -- cycle    ;
\draw    (224.56,162.22) -- (234.78,141.57) ;
\draw [shift={(235.67,139.78)}, rotate = 116.34] [fill={rgb, 255:red, 0; green, 0; blue, 0 }  ][line width=0.08]  [draw opacity=0] (12,-3) -- (0,0) -- (12,3) -- cycle    ;
\draw    (174.56,248.67) -- (187.45,229.22) ;
\draw [shift={(188.56,227.56)}, rotate = 123.55] [fill={rgb, 255:red, 0; green, 0; blue, 0 }  ][line width=0.08]  [draw opacity=0] (12,-3) -- (0,0) -- (12,3) -- cycle    ;
\draw    (208.64,191.82) -- (218.94,172.65) ;
\draw [shift={(219.89,170.89)}, rotate = 118.26] [fill={rgb, 255:red, 0; green, 0; blue, 0 }  ][line width=0.08]  [draw opacity=0] (12,-3) -- (0,0) -- (12,3) -- cycle    ;
\draw    (154.78,277.78) -- (168.72,258.51) ;
\draw [shift={(169.89,256.89)}, rotate = 125.88] [fill={rgb, 255:red, 0; green, 0; blue, 0 }  ][line width=0.08]  [draw opacity=0] (12,-3) -- (0,0) -- (12,3) -- cycle    ;
\draw    (296.64,197.09) -- (309.52,178.02) ;
\draw [shift={(310.64,176.36)}, rotate = 124.04] [fill={rgb, 255:red, 0; green, 0; blue, 0 }  ][line width=0.08]  [draw opacity=0] (12,-3) -- (0,0) -- (12,3) -- cycle    ;
\draw    (236.09,282.18) -- (249.93,264.84) ;
\draw [shift={(251.18,263.27)}, rotate = 128.59] [fill={rgb, 255:red, 0; green, 0; blue, 0 }  ][line width=0.08]  [draw opacity=0] (12,-3) -- (0,0) -- (12,3) -- cycle    ;
\draw    (331.73,138) -- (342.76,117.93) ;
\draw [shift={(343.73,116.18)}, rotate = 118.81] [fill={rgb, 255:red, 0; green, 0; blue, 0 }  ][line width=0.08]  [draw opacity=0] (12,-3) -- (0,0) -- (12,3) -- cycle    ;
\draw    (256.82,255.27) -- (270.35,237.06) ;
\draw [shift={(271.55,235.45)}, rotate = 126.62] [fill={rgb, 255:red, 0; green, 0; blue, 0 }  ][line width=0.08]  [draw opacity=0] (12,-3) -- (0,0) -- (12,3) -- cycle    ;
\draw    (314.36,168.73) -- (325.7,149.9) ;
\draw [shift={(326.73,148.18)}, rotate = 121.04] [fill={rgb, 255:red, 0; green, 0; blue, 0 }  ][line width=0.08]  [draw opacity=0] (12,-3) -- (0,0) -- (12,3) -- cycle    ;
\draw    (276.45,227.64) -- (289.13,209.46) ;
\draw [shift={(290.27,207.82)}, rotate = 124.89] [fill={rgb, 255:red, 0; green, 0; blue, 0 }  ][line width=0.08]  [draw opacity=0] (12,-3) -- (0,0) -- (12,3) -- cycle    ;
\draw    (346.48,108.29) -- (355.37,88.19) ;
\draw [shift={(356.18,86.36)}, rotate = 113.86] [fill={rgb, 255:red, 0; green, 0; blue, 0 }  ][line width=0.08]  [draw opacity=0] (12,-3) -- (0,0) -- (12,3) -- cycle    ;
\draw [color={rgb, 255:red, 0; green, 0; blue, 0 }  ,draw opacity=1 ][line width=1.5]    (99.28,165.95) .. controls (267,79.45) and (290,189.95) .. (459.67,98.42) ;
\draw [color={rgb, 255:red, 0; green, 0; blue, 0 }  ,draw opacity=1 ][line width=1.5]    (459.67,98.42) .. controls (475.45,89.95) and (476.45,90.45) .. (490.95,82.45) ;

\draw [color={rgb, 255:red, 0; green, 0; blue, 0 }  ,draw opacity=1 ][line width=1.5]    (98.57,214.45) .. controls (233.65,152.45) and (273.15,236.95) .. (415.18,167.95) ;
\draw [color={rgb, 255:red, 0; green, 0; blue, 0 }  ,draw opacity=1 ][line width=1.5]    (415.18,167.95) .. controls (437.65,156.45) and (471.15,139.45) .. (491.65,131.45) ;

\draw [color={rgb, 255:red, 0; green, 0; blue, 0 }  ,draw opacity=1 ][line width=1.5]    (98.57,263.95) .. controls (215.15,220.55) and (262.3,295.55) .. (399.99,222.79) ;
\draw [color={rgb, 255:red, 0; green, 0; blue, 0 }  ,draw opacity=1 ][line width=1.5]    (399.99,222.79) .. controls (420.3,213.05) and (464.45,189.55) .. (491.8,182.55) ;

\draw [line width=0.75]  (113,295.7) -- (491.8,295.7)(123.6,48.7) -- (123.6,306) (484.8,290.7) -- (491.8,295.7) -- (484.8,300.7) (118.6,55.7) -- (123.6,48.7) -- (128.6,55.7)  ;

\draw (498.5,174.5) node [anchor=north west][inner sep=0.75pt]  [font=\small]  {$y_{1}( t)$};
\draw (498.5,121.9) node [anchor=north west][inner sep=0.75pt]  [font=\small]  {$y_{2}( t)$};
\draw (498.5,72.9) node [anchor=north west][inner sep=0.75pt]  [font=\small]  {$y_{3}( t)$};
\draw (481.6,260) node [anchor=north west][inner sep=0.75pt]  [font=\small]  {$t$};
\draw (133.07,47.67) node [anchor=north west][inner sep=0.75pt]  [font=\small]  {$y$};
\draw (359.78,87.36) node [anchor=north west][inner sep=0.75pt]  [font=\small]  {$( \xi ,\eta )$};
\end{tikzpicture}
\caption{Generator of a Lie symmetry as a tangent vector field. $y_1$, $y_2$ and $y_3$ are solutions of the ODE for different initial values. The dotted lines represent the Lie symmetry, transforming the solutions into each other.}
\label{fig:generator_vector_field}
\end{figure}
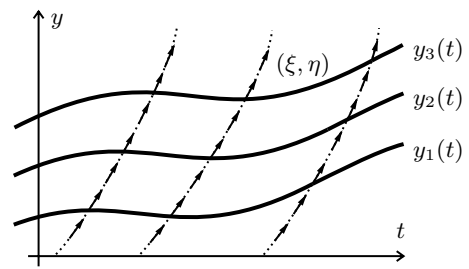

Instead of learning individual symmetry transformations, we would like to learn the whole one-parameter Lie group at once. It turns out that this is possible, because the Lie group, that is, all its symmetry transformations, is determined by its generator, which is the tangent vector field $X$ on $\mathbb{R}^{d+1}$ of the flow defined by $\theta \mapsto \{h_\theta(t,y)\}$ at $\theta =0$. The generator is unique up to scaling. Using the inner product $\langle \cdot,\cdot\rangle$ on $\mathbb{R}^d$ and $\nabla_y$, the gradient with respect to $y$, its components can be expressed as
\begin{align*}
    &X(t,y) = \xi(t,y)\partial_t + \big\langle \eta(t,y), \nabla_y \big\rangle \\
    &\text{with }\: \xi(t,y) \coloneqq \frac{\mathrm{d} \hat{t}}{\mathrm{d}\theta}\Big\vert_{\theta=0} 
    \:\text{and}\:\: \eta(t,y) \coloneqq \frac{\mathrm{d}\hat{y}}{\mathrm{d}\theta}\Big\vert_{\theta=0}. 
\end{align*}
The concept of a generator is visualized in Figure~\ref{fig:generator_vector_field}.

For learning the generator, it remains to express the symmetry condition $\hat{y}' = f(\hat t, \hat y)$ in terms of the functions $\xi$ and $\eta$. The symmetry condition states that the zero set of $\hat{y}' -f(\hat t, \hat y)$ does not change under symmetry transformations, which for one-parameter Lie groups can be expressed as 
\begin{align*}
0 &= \frac{\mathrm{d}}{\mathrm{d}\theta}\Big\vert_{\theta=0} \big(\hat{y}' - f(\hat{t},\hat{y})\big) \\
&= \frac{\mathrm{d}\hat{y}'}{\mathrm{d}\theta}\Big\vert_{\theta=0}\, - \frac{\partial f}{\partial \hat t} \frac{\mathrm{d}\hat{t}}{\mathrm{d} \theta}\Big\vert_{\theta=0}\, - \Big( \nabla_{\hat y} f \Big) \cdot \frac{\mathrm{d} \hat{y}}{\mathrm{d} \theta} \Big\vert_{\theta=0} \\
&= \frac{\mathrm{d}\hat{y}'}{\mathrm{d}\theta}\Big\vert_{\theta=0}\, - \frac{\partial f}{\partial t} \frac{\mathrm{d}\hat{t}}{\mathrm{d} \theta}\Big\vert_{\theta=0}\, - \Big( \nabla_y f \Big) \cdot \frac{\mathrm{d} \hat{y}}{\mathrm{d} \theta} \Big\vert_{\theta=0} \\
&= \frac{\mathrm{d}\hat{y}'}{\mathrm{d}\theta}\Big\vert_{\theta=0}\, - \frac{\partial f}{\partial t} \xi(t,y)\, - \Big( \nabla_y f \Big)\cdot \eta (t,y),
\end{align*}
where $\nabla_y f$ is a Jacobi matrix.  For the second to last equation, we have used that 
\[
\hat{t} (\theta =0, t, y) = t  \quad\mathrm{and}\quad\hat{y}(\theta =0, t, y) =y, 
\]
and the last equality follows from the definition of the generator $X$. The only expression in the symmetry condition that still involves $\hat t$ and $\hat y$ is $\frac{\mathrm{d}\hat{y}'}{\mathrm{d}\theta}\Big\vert_{\theta=0}$. For rewriting this expression as a function of $t$ and $y$, we expand $\hat{y}$ and $\hat{t}$ in orders of the parameter $\theta$,  
\begin{align*}
    \hat{y} &= y + \theta\cdot \eta(t,y) + o(\theta^2) \\
    \hat{t} &= t + \theta\cdot \xi(t,y) + o(\theta^2).
\end{align*}
From plugging the expansions into the derivative $\hat{y}'$,
\begin{align*}
    \hat{y}' &= \frac{\mathrm{d}\hat{y}}{\mathrm{d}\hat{t}} = \frac{\mathrm{d} \hat{y}}{\mathrm{d} t} \frac{\mathrm{d} t}{\mathrm{d}\hat{t}} = \frac{\mathrm{d} \hat{y}}{\mathrm{d} t} \Big(\frac{\mathrm{d} \hat{t}}{\mathrm{d} t}\Big)^{-1} = \frac{y'+\theta \frac{\mathrm{d}\eta}{\mathrm{d}t}+o(\theta^2)}{1+\theta\frac{\mathrm{d}\xi}{\mathrm{d}t}+o(\theta^2)} \\ 
    &= y' + \theta\left( \frac{\mathrm{d}\eta}{\mathrm{d}t} - y'\frac{\mathrm{d}\xi}{\mathrm{d}t} \right) + o(\theta^2),
\end{align*}
where the last equality is implied by 
\[
\left(1 + \theta a + o(\theta^2)\right)^{-1} = 1-\theta a + o(\theta^2),
\]
it follows that
\begin{align*}
    \frac{\mathrm{d}\hat{y}'}{\mathrm{d}\theta}\Big\vert_{\theta=0} &= \frac{\mathrm{d}\eta}{\mathrm{d}t} - y'\frac{\mathrm{d}\xi}{\mathrm{d}t} \\  
    &= \frac{\partial \eta}{\partial t} + \big( \nabla_{y} \eta\big)\cdot y' - \Big( \frac{\partial \xi}{\partial t} + \big\langle\nabla_y \xi, y'\big\rangle\Big)y'.
\end{align*}
Here, $\nabla_{y} \eta$ is also a Jacobi matrix. 

The symmetry condition can finally be expressed only in terms of functions of $t$ and $y$, namely, 
\[
\frac{\partial \eta}{\partial t} + \big( \nabla_{y} \eta\big)\cdot y' - \Big( \frac{\partial \xi}{\partial t} + \big\langle\nabla_y \xi, y'\big\rangle\Big)y' - \xi\frac{\partial f}{\partial t} -\big( \nabla_y f \big) \cdot \eta  = 0. 
\]

\section{Hamiltonian Symmetry}

In principle, the symmetry condition is all we need to define and compute a symmetry loss function that can guide our search for symmetries. 
Given $f$, we search for functions $\eta(t, y)$ and $\xi (t,y)$, that fulfill the symmetry condition.
In practice, however, there remain technical challenges when it comes to estimating a meaningful generator. Every ODE has at least one one-parameter symmetry group, namely, its Hamiltonian symmetry, that is, the time-evolution symmetry,
\[
h_\theta(t,y(t)) = \big(t+\theta,y(t+\theta)\big), 
\]
with symmetry generator
\[
X_H \coloneqq \partial_t + \langle f,\nabla_y\rangle.
\]
The Hamiltonian generator $X_H(t,y)$ and actually also any multiple $\xi (t,y)X_H(t,y)$ are indeed symmetry generators. This can be seen by plugging $\xi X_H$ into the symmetry condition, which gives
\begin{align*}
    \xi\left( \frac{\mathrm{d} f}{\mathrm{d} t} - \frac{\partial f}{\partial t} -\big( \nabla_y f \big) \cdot f \right)
    + f \frac{\mathrm{d} \xi}{\mathrm{d} t} - \frac{\mathrm{d} \xi}{\mathrm{d} t} y' = 0,
\end{align*}
where the terms cancel, because $y'=f$. However, the Hamiltonian symmetry $X_H$ and any multiple $\xi X_H$ do not provide more information than the already known function $f$ and thus can not be used for simplifying the ODE system.
Yet, a symmetry generator $X$ might contain a multiple of the Hamiltonian symmetry, 
\begin{align*}
X &= \xi \partial_t + \big\langle \eta, \nabla_y \big\rangle \\
&= \xi \partial_t + \xi \langle f,\nabla_y\rangle - \xi \langle f,\nabla y\rangle + \big\langle \eta, \nabla_y \big\rangle \\
&= \xi \big( \partial_t + \langle f,\nabla_y\rangle \big) + \big\langle \eta -\xi f,\nabla_y \big\rangle \\
&= \xi X_H + \big\langle \eta -\xi f,\nabla_y \big\rangle.
\end{align*}
Since the Hamiltonian symmetry as well as its multiples are not informative for our task, we remove the contribution of the Hamiltonian symmetry from any estimated symmetry generator, and only consider the non-trivial part
\[
\langle \eta^\star, \nabla_y \rangle \coloneqq \big\langle \eta -\xi f,\nabla_y \big\rangle = X - \xi X_H.
\]
The non-trivial part is also a generator, because generators form a vector space, $X$ is a generator by definition, and we have established already that $\xi X_H$ is a generator. It provides information about a transformation from one time series into another, but not the trivial information about traversing a solution (time-evolution), which is given by the Hamiltonian symmetry.

Removing the trivial part of the generator also simplifies the symmetry condition, which becomes, 
\begin{align}
0 = &\frac{\partial \eta^\star}{\partial t} + \big( \nabla_{y} \eta^\star\big)\cdot y' - \Big( \nabla_y f \Big) \cdot \eta^\star \nonumber \\ 
\label{eq:generator_condition}
&= \nabla \eta^\star \cdot \begin{bmatrix} 1 \\ f \end{bmatrix} - \nabla f \cdot \begin{bmatrix} 0 \\ \eta^\star \end{bmatrix}, 
\end{align}
where $\nabla \eta^\star$ and $\nabla f$ are Jacobi matrices of derivatives with respect to $t$ and $y$. 

By definition, any generator $X = \langle \eta^\star, \nabla_y \rangle$, that satisfies the simplified symmetry condition, does not contain a multiple of the Hamiltonian symmetry generator. Therefore, the search space of generators is reduced to the space of informative generators, that do not contain redundant information from the Hamiltonian generator. 

The implicit removal of the trivial part of symmetry generators also facilitates the simultaneous use of information from different symmetries, by considering multiple independent generators. A set $\{X_k\}_{k=1}^n$ of generators is called \emph{dependent} if all the generators $X_k$ are linearly dependent along all solutions, that is, there exist non-zero constants $\{c_k\}_{k=1}^n$ such that $\sum_{k=1}^n c_k X_k(t,y) = 0$ along all solutions of the ODE. Without implicitly (or explicitly) removing the trivial part of the generators, it is not straightforward to find independent generators that are informative. To see this, consider the generators $X=\langle \eta^\star, \nabla_y \rangle$ and $X'=X+\kappa X_H$, with a non-zero continuously differentiable function $\kappa(t,y)$. The two generators contain, for our purpose, the same information. They are, however, independent, as the $\partial_t$ component is zero for $X$ and non-zero for $X'$. This shows, that ignoring multiples of the Hamiltonian symmetry generator makes independence of generators a meaningful measure of independent information. 


\section{Symbolic Regression for Symmetry Finding}
\label{sec:generatorLoss}

\begin{figure}[h!]
    \centering
    \scalebox{0.83}{\input{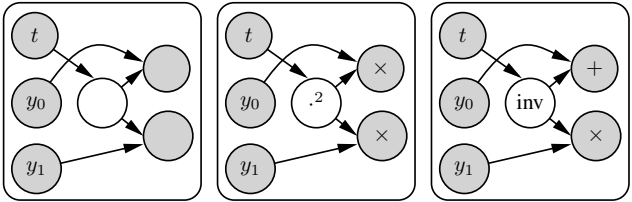}}
    \caption{The symbolic regression framework \citep{kahlmeyer2024udfs} works by creating DAG skeletons (left), that are turned into expression DAGs by assigning valid function symbols to the empty operator nodes (middle and right). Here, the assignment in the middle produces the expression $\eta^\star(t, y_1, y_2) = (t^2y_1, t^2y_2)$ and the assignment on the right produces the expression $\eta^\star(t, y_1, y_2) = (y_0+t^{-1}, y_1t^{-1})$.}
    \label{fig:dagsearch_explanation}
\end{figure}

Given a $d$-dimensional ODE system, our goal is use symbolic regression to search for a generator function 
\[
\eta^\star: \mathbb{R}^{d+1}\rightarrow\mathbb{R}^d
\]
in symbolic form, that satisfies the symmetry condition in Equation~\ref{eq:generator_condition}. The structure of the symmetry condition is such that it requires us to compute all outputs of $\eta^\star$ simultaneously, which is a problem, because most of the established symbolic regressors split multi-output regression problems into single-output regression problems. An exception is the state-of-the-art symbolic regression framework \citep{kahlmeyer2024udfs} that can directly deal with multi-output regression problems. \citet{kahlmeyer2024udfs} construct their regressor in two steps: First, expression DAG skeletons are created up to a given size that depends on the available computational budget. Here, DAG skeletons are expression DAGs whose operator nodes have not yet been labeled with function symbols. Second, each DAG skeleton is turned into a family of expression DAGs by validly labeling its operator nodes with function symbols. The expression DAGs are then scored by the given loss function, and the regressor is given by the expression DAG that incurs the smallest loss. The construction process is illustrated in Figure~\ref{fig:dagsearch_explanation}.

For symbolic regression, \citet{kahlmeyer2024udfs} use the mean-squared error (MSE) between a DAG's output and the given observations as loss function. For our task of finding symmetries of ODE systems, we need to replace this loss function by a loss function that is based on the symmetry condition. Such a loss function is constructed as follows: Let $f: \mathbb{R}^{d+1}\rightarrow\mathbb{R}^d$ be the symbolic form of the ODE system under investigation. We first solve the ODE numerically by the Runge-Kutta algorithm~\cite{rungekutta_1980} that computes a sample $\mathcal{D} = \big(t_i, y(t_i)\big)_{i=1,\dots,n}$ of the ODE system's solution. For each dimension $k=1,\dots, d$ of the ODE system, we can evaluate a candidate symmetry generator $\eta^\star$ at a given value pair $(t, y)\in\mathcal{D}$, which gives 
\[
     S_{\eta^\star}^k(t, y) = \nabla\eta^\star_k (t,y)\begin{bmatrix} 1 \\ \ f(t,y) \end{bmatrix} - \nabla f_k(t,y)\begin{bmatrix} 0 \\ \eta^\star(t,y) \end{bmatrix}.
\]
The loss of the candidate symmetry generator $\eta^\star$ is then given as
\[
    L_{\mathcal{D},f}(\eta^\star) = \cfrac{1}{nd}\sum_{(t, y)\in\mathcal{D}}\sum_{k=1}^d \left(S_{\eta^\star}^k(t, y)\right)^2.
\]
The loss function evaluates to zero if, and only if the symmetry condition is satisfied at all points in the numeric solution  $\mathcal{D} = \big(t_i, y(t_i)\big)_{i=1,\dots,n}$ in every dimension of the ODE system. For implementing the loss function we need to compute the gradients $\nabla\eta^\star$. The internal representation of the DAG facilitates calculating these gradients via automatic differentiation, for instance, by the autodiff engine provided in PyTorch~\cite{paszke2017_pytorch}. 

Finally, note that the generator $\eta^\star = 0$ always satisfies the generator condition. The corresponding symmetry however gives no additional information about the system, as it maps every solution back onto itself.
In order to avoid the trivial generator, our implementation discards candidate expression DAGs for $\eta^\star$ where the \textit{median absolute value} of $\eta^\star$ evaluated at the data points is smaller than a threshold value of $\varepsilon = 0.01$. 

\begin{figure}[h!]
    \centering
    \tikzset{every picture/.style={line width=0.75pt}} 

\begin{tikzpicture}[x=0.75pt,y=0.75pt,yscale=-1,xscale=1]

\draw    (220,110.75) .. controls (285.17,116.25) and (275.17,73.25) .. (300,80.75) ;
\draw   (10,22.75) .. controls (10,16.12) and (15.37,10.75) .. (22,10.75) -- (118,10.75) .. controls (124.63,10.75) and (130,16.12) .. (130,22.75) -- (130,58.75) .. controls (130,65.38) and (124.63,70.75) .. (118,70.75) -- (22,70.75) .. controls (15.37,70.75) and (10,65.38) .. (10,58.75) -- cycle ;
\draw    (130,40.75) -- (188,40.75) ;
\draw [shift={(190,40.75)}, rotate = 180] [fill={rgb, 255:red, 0; green, 0; blue, 0 }  ][line width=0.08]  [draw opacity=0] (12,-3) -- (0,0) -- (12,3) -- cycle    ;
\draw   (191.25,33.75) .. controls (191.25,20.63) and (201.88,10) .. (215,10) -- (286.25,10) .. controls (299.37,10) and (310,20.63) .. (310,33.75) -- (310,127) .. controls (310,140.12) and (299.37,150.75) .. (286.25,150.75) -- (215,150.75) .. controls (201.88,150.75) and (191.25,140.12) .. (191.25,127) -- cycle ;
\draw    (220,130.75) -- (298,130.75) ;
\draw [shift={(300,130.75)}, rotate = 180] [color={rgb, 255:red, 0; green, 0; blue, 0 }  ][line width=0.75]    (10.93,-3.29) .. controls (6.95,-1.4) and (3.31,-0.3) .. (0,0) .. controls (3.31,0.3) and (6.95,1.4) .. (10.93,3.29)   ;
\draw    (220,130.75) -- (220.24,62.75) ;
\draw [shift={(220.25,60.75)}, rotate = 90.2] [color={rgb, 255:red, 0; green, 0; blue, 0 }  ][line width=0.75]    (10.93,-3.29) .. controls (6.95,-1.4) and (3.31,-0.3) .. (0,0) .. controls (3.31,0.3) and (6.95,1.4) .. (10.93,3.29)   ;
\draw  [fill={rgb, 255:red, 31; green, 119; blue, 180 }  ,fill opacity=1 ] (248.5,107.5) .. controls (248.5,104.32) and (251.07,101.75) .. (254.25,101.75) .. controls (257.43,101.75) and (260,104.32) .. (260,107.5) .. controls (260,110.68) and (257.43,113.25) .. (254.25,113.25) .. controls (251.07,113.25) and (248.5,110.68) .. (248.5,107.5) -- cycle ;
\draw  [fill={rgb, 255:red, 31; green, 119; blue, 180 }  ,fill opacity=1 ] (265.5,97.5) .. controls (265.5,94.32) and (268.07,91.75) .. (271.25,91.75) .. controls (274.43,91.75) and (277,94.32) .. (277,97.5) .. controls (277,100.68) and (274.43,103.25) .. (271.25,103.25) .. controls (268.07,103.25) and (265.5,100.68) .. (265.5,97.5) -- cycle ;
\draw  [fill={rgb, 255:red, 31; green, 119; blue, 180 }  ,fill opacity=1 ] (228.5,110.5) .. controls (228.5,107.32) and (231.07,104.75) .. (234.25,104.75) .. controls (237.43,104.75) and (240,107.32) .. (240,110.5) .. controls (240,113.68) and (237.43,116.25) .. (234.25,116.25) .. controls (231.07,116.25) and (228.5,113.68) .. (228.5,110.5) -- cycle ;
\draw  [fill={rgb, 255:red, 31; green, 119; blue, 180 }  ,fill opacity=1 ] (279.5,86.5) .. controls (279.5,83.32) and (282.07,80.75) .. (285.25,80.75) .. controls (288.43,80.75) and (291,83.32) .. (291,86.5) .. controls (291,89.68) and (288.43,92.25) .. (285.25,92.25) .. controls (282.07,92.25) and (279.5,89.68) .. (279.5,86.5) -- cycle ;
\draw   (190,182.75) .. controls (190,176.12) and (195.37,170.75) .. (202,170.75) -- (298,170.75) .. controls (304.63,170.75) and (310,176.12) .. (310,182.75) -- (310,218.75) .. controls (310,225.38) and (304.63,230.75) .. (298,230.75) -- (202,230.75) .. controls (195.37,230.75) and (190,225.38) .. (190,218.75) -- cycle ;
\draw    (250,150.75) -- (250,168.75) ;
\draw [shift={(250,170.75)}, rotate = 270] [fill={rgb, 255:red, 0; green, 0; blue, 0 }  ][line width=0.08]  [draw opacity=0] (12,-3) -- (0,0) -- (12,3) -- cycle    ;
\draw    (190,199.75) -- (132,199.75) ;
\draw [shift={(130,199.75)}, rotate = 360] [fill={rgb, 255:red, 0; green, 0; blue, 0 }  ][line width=0.08]  [draw opacity=0] (12,-3) -- (0,0) -- (12,3) -- cycle    ;
\draw   (10,182.15) .. controls (10,175.44) and (15.44,170) .. (22.15,170) -- (117.85,170) .. controls (124.56,170) and (130,175.44) .. (130,182.15) -- (130,218.6) .. controls (130,225.31) and (124.56,230.75) .. (117.85,230.75) -- (22.15,230.75) .. controls (15.44,230.75) and (10,225.31) .. (10,218.6) -- cycle ;
\draw    (70,70.75) -- (200.41,169.54) ;
\draw [shift={(202,170.75)}, rotate = 217.15] [fill={rgb, 255:red, 0; green, 0; blue, 0 }  ][line width=0.08]  [draw opacity=0] (12,-3) -- (0,0) -- (12,3) -- cycle    ;

\draw (65.5,58.35) node [anchor=south] [inner sep=0.75pt]    {$\dot{y} \ =\ f( t,\ y)$};
\draw (68,12.75) node [anchor=north] [inner sep=0.75pt]   [align=left] {ODE};
\draw (261,133.15) node [anchor=north west][inner sep=0.75pt]    {$t$};
\draw (201,83.15) node [anchor=north west][inner sep=0.75pt]    {$y$};
\draw (252,33.15) node [anchor=north] [inner sep=0.75pt]    {$\mathcal{D} \ =\ [ t_{i} ,y( t_{i})]$};
\draw (253.5,206.5) node    {$L_{\mathcal{D} ,\ f}\left( \eta ^{\star }\right)$};
\draw (65.5,171) node [anchor=north] [inner sep=0.75pt]   [align=left] {Generator};
\draw (251.5,181.25) node   [align=left] {Loss Function};
\draw (132,43.75) node [anchor=north west][inner sep=0.75pt]   [align=left] {RK45};
\draw (146,201.75) node [anchor=north west][inner sep=0.75pt]   [align=left] {SR};
\draw (248.5,12.75) node [anchor=north] [inner sep=0.75pt]   [align=left] {Time Series};
\draw (127,93.15) node [anchor=north west][inner sep=0.75pt]    {$f$};
\draw (15,200) node [anchor=north west][inner sep=0.75pt]    {$\text{argmin}_{\eta ^{\star }} \ L_{\mathcal{D} ,\ f}\left( \eta ^{\star }\right)$};

\end{tikzpicture}
    \caption{Flowchart of our method: From a given ODE, time series data $\mathcal{D}$ are simulated using the Runge-Kutta algorithm (RK45). Based on this data and $f$, a loss function for expressions is derived. We finally use a symbolic regression framework (SR)  to search for a symbolic symmetry generator $\eta^\star$ that minimizes the loss function.}
    \label{fig:flowchart}
\end{figure}
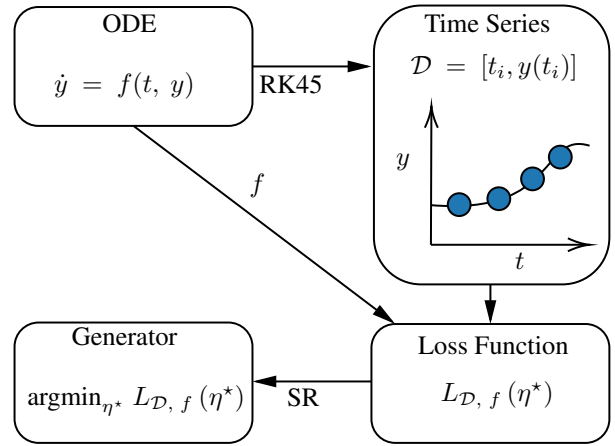

The whole procedure for finding a symbolic generator $\eta^\star$ is summarized in Figure~\ref{fig:flowchart}.

\section{Discussion}

We provide examples on which our symbolic regression method finds symmetry generators while the established algebraic symmetry finding methods struggle to do so. That is, the symbolic regression approach enlarges the set of problems that can be tackled by symmetry finding methods. 
Here, we compare our approach only to Mathematica's well-maintained SYM package~\citep{Stylianos2005_SYM}, because it essentially subsumes the other packages. Maple's symgen package cannot handle ODE systems, and the MathLie package heavily relies on ODE-specific user interaction.
In the following, we first use the example ODE system from the introduction to illustrate different modes that are used by SYM for finding symmetries, before we provide example problems that are difficult for SYM. Finally, we briefly discuss the added benefit of searching for independent generators and provide running times for the search-based symbolic regression approach.

\subsection{SYM's Symmetry Finding Approach}

We illustrate SYM's symmetry finding approach, that comprises an automatic and an interactive mode, on the simple ODE system from the introduction,
\[
f(t, y_1, y_2) = 
\begin{bmatrix}
-y_2 \\ y_1
\end{bmatrix}.
\]
We are looking for a generator 
\begin{align*}
    X(t, y_1, y_2) = \left\langle\begin{bmatrix}
        \xi(t, y_1, y_2)\\
        \eta_1(t, y_1, y_2)\\
        \eta_2(t, y_1, y_2)\\
    \end{bmatrix}, \begin{bmatrix}
        \partial_t\\
        \partial_{y_1}\\
        \partial_{y_2}
    \end{bmatrix}\right\rangle
\end{align*}
in the form of the functions $\xi$ and $\eta_i$.
Note that, since this ODE system can be solved directly by Mathematica, there is no need for a symmetry analysis when it comes to solving the system. The system is useful, however, for demonstrating SYM's algebraic symmetry finding approach.

\paragraph{Automatic mode.}

Given the symbolic form of $f$ in our example system, the SYM package derives the PDEs for the general generator condition,
\begin{align*}
0&=&\eta_2+y_1\frac{\partial\eta_1}{\partial y_2}+y_1y_2\frac{\partial\xi}{\partial y_2}-y_2\frac{\partial\eta_1}{\partial y_1}\\
&&-y_2^2\frac{\partial\xi}{\partial y_1}+\frac{\partial\eta_1}{\partial t} + y_2\frac{\partial \xi}{\partial t}\\
0&=&\eta_1-y_1\frac{\partial\eta_2}{\partial y_2}+y_1^2\frac{\partial\xi}{\partial y_2} + y_2\frac{\partial\eta_2}{\partial y_1}\\
&&-y_1y_2\frac{\partial\xi}{\partial y_1} - \frac{\partial\eta_2}{\partial t} + y_1\frac{\partial\xi}{\partial t}\\
\end{align*}
with the unknown functions $\xi$ and $\eta_i$.
SYM then iteratively attempts to reduce the number of PDEs by solving one PDE for an unknown function and substituting the resulting expression into the remaining PDEs until only one PDE remains. The remaining PDE acts as a constraint for the solutions that were used as substitutions.
In our example, SYM solves the second PDE for $\eta_1$, that is,
\begin{align*}
\eta_1 = &\quad\: y_1\frac{\partial\eta_2}{\partial y_2}-y_1^2\frac{\partial\xi}{\partial y_2} - y_2\frac{\partial\eta_2}{\partial y_1}\\
 &+y_1y_2\frac{\partial\xi}{\partial y_1} + \frac{\partial\eta_2}{\partial t} - y_1\frac{\partial\xi}{\partial t},
\end{align*}
that is plugged into the first equation, which then acts as a constraint. 
For finding generators, one now has to solve the PDE constraint and use the solution to calculate $\eta_1$.
In general, we consider this approach successful if all PDE constraints can be solved by Mathematica.
This is, however, already not the case for our simple example system.

\paragraph{Interactive mode.}

Alternatively, the user can \textit{interactively} fix all but one of the unknown functions of the PDE constraints and then use Mathematica to solve for the remaining functions. 
For our example system, when setting $\eta_2 = \xi = 0$ in order to exclude the Hamiltonian symmetry, the first PDE reduces to
\begin{align*}
0&=y_1\frac{\partial\eta_1}{\partial y_2}-y_2\frac{\partial\eta_1}{\partial y_1}+\frac{\partial\eta_1}{\partial t},
\end{align*}
which Mathematica can solve with the result
\begin{align*}
\eta_1 = y_1\cos(t) + y_2 \sin(t).
\end{align*}
Our search-based symbolic regression approach implicitly excludes the Hamiltonian symmetry by setting $\xi = 0$ and directly finds the generators $\eta_1 = \cos(t)$ and $\eta_2 = \sin(t)$, which can be verified to satisfy the generator condition.

\subsection{Hard Problems for Mathematica and SYM}

We demonstrate the added value of the symbolic regression approach on ten representative ODE systems for which Mathematica does not find an explicit solution directly, that is, without the SYM package. 
More details on the ODEs can be found in the supplement.

We test SYM's symmetry finding abilities on the ten examples by running it in three modes:
First, the automatic mode, where SYM's result is a closed form expression for one of the generator parts, with constraints. The constraints are themselves a system of PDEs. We consider SYM successful, if the constraint PDEs can be solved by Mathematica.
The second and third approach are variants of the interactive mode. We either set all but one component to zero (SYM-Z) or to the values of a known generator (SYM-S) and solve for the missing components. Note, however, that the latter approach is usually infeasible in practice. The results are shown in Table~\ref{tab:results}.

\newcommand\tabspace{0.45cm}
\newcommand{\greencheck}{{\color{LimeGreen}\cmark}}
\newcommand{\redcross}{{\color{BrickRed}\xmark}}
\begin{table}[ht]
    \centering
    \begin{tabular}{llccc}
    \multicolumn{2}{c}{ODE: $f(y_1,y_2) =$}&SYM&SYM-Z&SYM-S\\
    \toprule
    1&$\begin{bmatrix}
    y_{1}(t + y_{2}y_{1}^{-1})^2\\
    t^2y_{1}
    \end{bmatrix}$&\redcross&\redcross&\greencheck\\[\tabspace]
    2&$\begin{bmatrix}
    y_{1}^2y_{2}\exp(-y_{1})\\
    t\exp\left(-\frac{1}{y_{1}}\right)
    \end{bmatrix}$&\redcross&\redcross&\greencheck\\[\tabspace]
    3&$\begin{bmatrix}
    ty_{1}(y_{2} - \ln(y_{1}))\\
    t + y_{2} - \ln(y_{1})
    \end{bmatrix}$&\redcross&\redcross&\greencheck\\[\tabspace]
    4&$\begin{bmatrix}
    \frac{2y_{1} + y_{2}\exp(-\frac{y_{1}}{t^{2}})}{t}\\ 
    y_{2}
    \end{bmatrix}$&\redcross&\greencheck&\greencheck\\[\tabspace]
    5&$\begin{bmatrix}
    y_{1}(t - \ln(y_{1})\tan(t))\\
    y_{2}(1 -\ln(y_{1})\tan(t))
    \end{bmatrix}$&\redcross&\redcross&\greencheck\\[\tabspace]
    6&$\begin{bmatrix}
    y_{1}(ty_{2}y_{1}^{-1} + \frac{2\ln(y_{1})}{t})\\
    \frac{2y_{2}\ln(y_{1})}{t}
    \end{bmatrix}$&\redcross&\redcross&\greencheck\\[\tabspace]
    7&$\begin{bmatrix}
    \frac{y_{2}}{y_{1}}\exp(-\frac{y_{1}^2}{2t^{2}}) + \frac{y_{1}}{2t}\\
    -\frac{1}{2t^3}y_{1}^2y_{2}
    \end{bmatrix}$&\redcross&\redcross&\greencheck\\[\tabspace]
    8&$\begin{bmatrix}
    \sin(y_{2})\exp(-t)\\
    \sin(y_{1})\exp(-t)
    \end{bmatrix}$&\redcross&\redcross&\redcross\\[\tabspace]
    9&$\begin{bmatrix}
    ty_{2}\sin(y_{1})\\
    t\sin(y_{1})
    \end{bmatrix}$&\redcross&\redcross&\redcross\\[\tabspace]
    10&$\begin{bmatrix}
    t\ln(y_{2})\\
    ty_{1}^2
    \end{bmatrix}$&\redcross&\redcross&\redcross\\
    \end{tabular}
    \caption{Performance, indicated by success (\greencheck) and failure (\redcross), of SYM on ten representative example problems for which symbolic regression finds simple symmetry generators.}
    \label{tab:results}
\end{table}

The automatic procedure of SYM fails on all the examples, that is, not all the PDE constraints of the reformulated symmetry generator condition can be solved by Mathematica. 
When setting all but one generator function to zero, Mathematica is able to solve the determining equations only in one case. 
Even in the unrealistic case that all but one generator function are known, Mathematica fails to find a generator in three cases.

Our search-based symbolic regression approach automatically finds symmetry generators for all ten problems.

\subsection{Independent Generators}

We have argued that the implicit exclusion of the Hamiltonian symmetry allows searching for independent generators.
Here, we demonstrate this feature on the following nonlinear system,
\begin{align*}
f(t, y_1, y_2) &= \begin{bmatrix}
\sqrt{y_1}t\\
y_1y_2t
\end{bmatrix},
\end{align*}
where our approach discovers the two independent generators
\[
\eta^\star_1 =
\begin{bmatrix}
0 \\ y_2
\end{bmatrix}
\quad\textrm{ and }\quad
\eta^\star_2 =
\begin{bmatrix}
\sqrt{y_1} \\ y_1y_2
\end{bmatrix}.
\]
From the two independent generators we can derive two different coordinate transformations leading to two distinct one-dimensional ODEs,
\[
s_1'(r) = \sqrt{s_1}r
\quad\textrm{ and }\quad
s_2'(r) = 0.
\]
When either of  these systems can be solved, the solution can be translated back into the original coordinates, thus solving the original problem.
The trivial second ODE is solved by $s_2(r) = 0$ and leads to a correct solution,
\[
y_1(t) = \frac{t^4}{16}
\quad\textrm{ and }\quad
y_2(t) = e^{\frac{t^6}{96}},
\]
of the original problem.
Solving the first ODE with $s_1(r) = \frac{r^4}{16}$ leads to the same solution.
A detailed derivation of these results can be found in the supplementary material.

\subsection{Running Times}

The running time of the proposed method exclusively depends on the underlying symbolic regressor. By design, the running time of the regressor grows with the size of the search space for symbolic symmetry generators, which itself grows exponentially with the size of the covered expression DAGs~\citep{kahlmeyer2024udfs}. 

\begin{figure}[h!]
    \centering
    \includegraphics[width = \columnwidth]{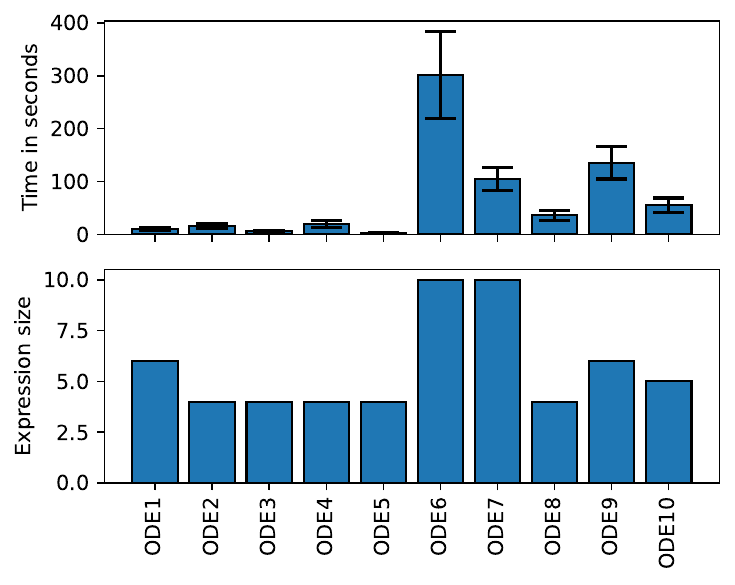}
    \caption{Top: Average running times and 95\% confidence intervals over 50 independent runs of the symbolic regression approach on the ten problems that are hard for SYM. 
    Bottom: Tree sizes of the \texttt{sympy} expressions~\citep{SymPy} for the symmetry generators found by the symbolic regressor.}
    \label{fig:runtimes_sizes}
\end{figure}

In Figure~\ref{fig:runtimes_sizes}, we show the average runtime for our constructed example problems together with the size of the expression tree of the found generators. As expected, our method is significantly faster when the generator has a simple expression. Nevertheless, even for the more challenging examples, generators are found within a reasonable time of under six minutes, running Python 3.10 on an Intel Xeon Gold 6226R 64-core processor with 128 GB of RAM.


\section{Conclusion}

We have introduced a novel application of symbolic regression, namely, the identification of symmetries of systems of ordinary, differential equations. Our search-based symbolic regression approach can automatically detect symmetries, that are difficult to find for traditional computer algebra systems. As a consequence, the search-based symbolic regression approach enlarges the space of ordinary differential equations, especially nonlinear differential equations, whose analytical solution can be approached using symmetry finding methods. 

\section*{Acknowledgements}
This work was supported by the Carl Zeiss Stiftung within the project "Interactive Inference".

\bibliography{aaai25}

\clearpage
\onecolumn
\appendix
\noindent
\rule{\textwidth}{0.3pt}
\begin{center}
  \textbf{\LARGE Discovering Symmetries of ODEs using Symbolic Regression (Supplementary Material)}
\end{center}
\textbf{\hfill submitted to AAAI 25}
\rule{\textwidth}{0.3pt}

\section{Reducing the Order of ODEs using Symmetries}




If a symmetry generator $X$ of an ODE system $y' = f(t,y), \, y\in \mathbb{R}^n$ is known, it can be used to find a set of coordinates in which this symmetry becomes a simple translation. Those coordinates are called canonical coordinates and can be determined by solving the first order PDEs
\begin{align*}
    X\cdot r &= 0, \\
    X\cdot v &= 1, \\
    X\cdot s_i &= 0, \text{ for } i\in\{1,\ldots,n-1\},
\end{align*}
where $(t,y)\to (r,v,s)$ is a coordinate transform (invertible and continuously differentiable in some domain).
The ODE in its original and canonical coordinates is invariant under the symmetry. The canonical coordinates are chosen in a way, that in these coordinates the symmetry function acts as a simple translation on the $v$ coordinate while leaving the other dimensions as they are:
\begin{alignat*}{3}
h_\theta(r) &= e^{\theta X}(r) =(1)(r) &&= r\\
h_\theta(v) &= e^{\theta X}(v) = (1 + \theta X)(v) &&= v +\theta\\
h_\theta(s_i) &= e^{\theta X}(s_i) =(1)(s_i) &&= s_i\\
\end{alignat*}
For $r$ and $s_i$ we used the fact, that only the first summand of the exponential series is non-zero. For $v$, the first two summands are non-zero, due to the definition of $r, v$ and $s_i$.

Canonical coordinates can be used to simplify the ODE into an ODE with dimension reduced by one and an integral. 
If we express the ODE in canonical coordinates
\begin{align*}
    \frac{\mathrm{d}v}{\mathrm{d}r} &= \frac{\frac{\partial v}{\partial t} + \langle \nabla_y v, y' \rangle}{\frac{\partial r}{\partial t} + \langle \nabla_y r, y' \rangle} = F_v(r,v,s), \\
    \frac{\mathrm{d}s_i}{\mathrm{d}r} &= \frac{\frac{\partial s_i}{\partial t} + \langle \nabla_y s_i, y' \rangle}{\frac{\partial r}{\partial t} + \langle \nabla_y r, y' \rangle} = F_i(r,v,s)\,,
\end{align*}
we can argue, that $F_v$ and $F_i$ do not depend on $v$.
The symmetry $h_\theta$ does also apply to the ODE in canonical coordinates. This means for $\hat r = h_\theta(r), \hat v = h_\theta(v) , \hat s_i = h_\theta(s_i)$ we have that
\begin{align*}
F_v(r, v, s_i) =\dv{v}{r} =\dv{(v + \theta)}{r}=\dv{\hat v}{\hat r}=F_v(\hat r, \hat v, \hat s_i)=F_v(r, v+\theta, s_i)\,.
\end{align*}
Similar for $F_i$:
\begin{align*}
F_i(r, v, s_i) =\dv{s_i}{r} =\dv{\hat s_i}{\hat r}=F_i(\hat r, \hat v, \hat s_i)=F_i(r, v+\theta, s_i)\,.
\end{align*}
In other words, both $F_v$ and $F_i$ do not change, if we put in arbitrary values for $\theta$ and are thus independent of $v$.
Hence, the ODE in canonical parameters can be written as
\begin{align*}
		\dv{v}{r} &= F_v(r, s)\\
		\dv{s_i}{r} &= F_i(r, s)\,,
\end{align*} 
where $F_i$ and $F_v$ do not depend on $v$ anymore. 

Solving this system means finding functions $s(r)$ and $v(r)$ that satisfy the system.\\
Since $\dv{s_i}{r}$ does not depend on $v$, it forms a system of ODEs of dimension $n-1$, that is one less than the original system.
Given the solution $s(r)$ of this ODE system, we can calculate $v(r)$ by integrating both sides of the equation for $\dv{v}{r}$:
\begin{equation*}
    v(r) = \int_r F_v(\tilde{r},s(\tilde{r})) \mathrm{d}\tilde{r} + \text{const.}
\end{equation*}
Note that this is only possible, because $F_v$ does not depend on $v$ anymore.

This means a solution of the original ODE can be inferred from a solution of the system above with reduced dimension.
The following two examples for this procedure have been used in the paper.

\subsection{Example: Introduction}
The 2D system of ODEs:
\begin{align*}
    y_1' &= -y_2\\
    y_2' &= y_1\,,
\end{align*}
together with a known and simple generator
\begin{align*}
X &= y_1\partial_{y_1} + y_2\partial_{y_2}\,,
\end{align*}
leads to the following functions $r, v, s$ that fullfill the PDE conditions:

\begin{center}
\begin{tabular}{lll}
Condition&Transformation&Inverse\\
\toprule
$X\cdot r = 0$&$r=t$&$t=r$\\
$X\cdot v = 1$&$v=\ln(y_1)+y_2y_1^{-1}$&$y_1=e^{v-s}$\\
$X\cdot s = 0$&$s=y_2y_1^{-1}$&$y_2=e^{v-s}s$\\
\end{tabular}
\end{center}
The ODE in these new coordinates becomes
\begin{align*}
\dv{v}{r}&=\cfrac{y_1^2-y_1y_2+y_2^2}{y_1^2}\\
&=1-s+s^2\\
\dv{s}{r}&=1+\cfrac{y_2^2}{y_1^2}\\
&=1+s^2\\
\end{align*}
The solution to $s'(r) = 1+s^2$ is the tangent function 
\begin{align*}
s(r) = \tan(r)\,.
\end{align*}
We then get $v(r)$ by calculating the integral\\
\begin{align*}
v(r) &= \int_r 1-\tan(\hat r) +\tan(\hat r)^2 d\hat r\\
&=\tan(r) + \log(\cos(r))\,.
\end{align*}
Using the inverse of the coordinate transformation, we finally get the solution in the original coordinates:
\begin{align*}
t&=r\\
y_1(t)&=\cos(t)\\
y_2(t)&=\cos(t)\tan(t)=\sin(t)
\end{align*}

\subsection{Example: Independent Generators}
Our proposed method enables us to search for multiple, independent generators. As an example we used the nonlinear 2D system of ODEs:
\begin{align*}
    y_1' &= \sqrt{y_1}t\\
    y_2' &= y_1y_2t\,,
\end{align*}
where our method finds the two independent generators
\begin{align*}
X_0 &= y_2\partial_{y_2}\\
X_1 &= \sqrt{y_1}\partial_{y_1} + y_1y_2\partial_{y_2}\,.
\end{align*}
These generators lead to the following functions $r, v, s$ that fullfill the PDE conditions:
\begin{center}
\begin{tabular}{lll}
Condition&Transformation&Inverse\\
\toprule
\multicolumn{3}{c}{$X_0$}\\
\midrule
$X\cdot r = 0$&$r=t$&$t=r$\\
$X\cdot v = 1$&$v=\log(y_2)$&$y_2=\exp(v)$\\
$X\cdot s = 0$&$s=y_1$&$y_1=s$\\
\midrule
\multicolumn{3}{c}{$X_1$}\\
\midrule
$X\cdot r = 0$&$r=t$&$t=r$\\
$X\cdot v = 1$&$v=2\sqrt{y_1}$&$y_1=(\frac{v}{2})^2$\\
$X\cdot s = 0$&$s=\frac{2}{3}\sqrt{y_1^3} - \log(y_2)$&$y_2=\exp\left(\frac{v^3}{12}-s\right)$\\
\end{tabular}
\end{center}

\noindent
For $X_0$, the ODE in these new coordinates becomes
\begin{align*}
\dv{v}{r}&=sr\\
\dv{s}{r}&=\sqrt{s}r\,.
\end{align*}
A solution for $s'(r) = \sqrt{s}r$ would be the polynomial function
\begin{align*}
s(r) = \frac{r^4}{16}\,,
\end{align*}
and $v(r)$ is found by calculating the integral
\begin{align*}
v(r) &= \int_r s(\hat r)\hat r d\hat r\\
&= \frac{1}{16}\int_r \hat r^5 d\hat r\\
&= \frac{r^6}{96}\,.
\end{align*}
Applying the inverse coordinate transformation of $X_0$, we get
\begin{align*}
t&=r\\
y_1&=\frac{t^4}{16}\\
y_2&=\exp\left(\frac{t^6}{96}\right)\,.
\end{align*}

\vspace{5pt}
\hrule
\vspace{6pt}

\noindent
For $X_1$, the ODE in these new coordinates becomes
\begin{align*}
\dv{v}{r}&=r\\
\dv{s}{r}&=0\,.
\end{align*}
A solution for $s'(r) = 0$ would be the constant function
\begin{align*}
s(r) = 0\,,
\end{align*}
and $v(r)$ is found by calculating the integral
\begin{align*}
v(r) &= \int_r \hat r d\hat r\\
&= 0.5r^2\,.
\end{align*}
Applying the inverse coordinate transformation of $X_1$, we get
\begin{align*}
t&=r\\
y_1&=\frac{t^4}{16}\\
y_2&=\exp\left(\frac{t^6}{96}\right)\,.
\end{align*}

\vspace{5pt}
\hrule
\vspace{6pt}

Both generators lead to the same solution of the target system. Nevertheless, there is an inherent tradeoff between the difficulty of finding the coordinate transformation and solving the reduced system. While for $X_1$ the reduced system is trivial, its coordinate transformation is more difficult compared to $X_0$. $X_0$ on the other hand has a simple coordinate transformation with a fairly simple reduced system (although not as trivial as for $X_1$).

\section{Symmetries of Higher Order ODEs}
In the main part of the paper we gave a short introduction to Lie point symmetries of ODEs. This section is dedicated to higher order ODEs and showing how their symmetry conditions are satisfied by considering the corresponding first order system of ODEs. We first need to introduce a concept which was implicitly used in the main part of the paper, so called prolongations of a generator.

\subsection{Prolongations of a Generator}
\label{subsec:Prolongations}
The calculations in this subsection can be found in e.g. \cite{stephani93} or \cite{bluman89}.\\ 
In order to describe the transformation of ODEs containing higher derivatives of $y$ we need to prolong a generator $X$ to the tangent spaces corresponding to these variables.
In the main part we already saw that the derivative $y'$ transforms under the Lie symmetry according to
\begin{equation*}
    \hat y' = y' + \theta \cdot\left( \frac{\mathrm{d}\eta}{\mathrm{d}t} - y'\frac{\mathrm{d}\xi}{\mathrm{d}t} \right) + o(\theta^2).
\end{equation*}
This means the prolonged generator $X^{(1)}$ should act on the derivative via
\begin{align*}
    X^{(1)}\cdot y' = \frac{\mathrm{d}\hat{y}'}{\mathrm{d}\theta}\Big\vert_{\theta=0} = \frac{\mathrm{d}\eta}{\mathrm{d}t} - y'\frac{\mathrm{d}\xi}{\mathrm{d}t} \eqqcolon \eta^{(1)}.
\end{align*}
Which results in the prolonged generator being $X^{(1)} = X + \eta^{(1)}_i \partial_{y'_i}$\\
The $n$-th prolongation 
\begin{align}
    X^{(n)} &\coloneqq X^{(n-1)} + \eta^{(n)}_i \partial_{y^{(n)}_i} \nonumber \\
    \text{with } \eta^{(n)} &\coloneqq \frac{\mathrm{d}\eta^{(n-1)}}{\mathrm{d}t} - y'\frac{\mathrm{d}\xi}{\mathrm{d}t} \label{eq: prolongation higher order}
\end{align}
acting on derivatives of $y$ up to the $n$-th order can be derived in the same way.

\subsection{Higher Order ODEs}
Every higher order ODE $\frac{\mathrm{d}^ny}{\mathrm{d}t^n} = y^{(n)}=f(t,y,\ldots,y^{(n-1)})$ can be expressed via an equivalent first order ODE by introducing new variables $z_k = y^{(k)}$:
\begin{align}
    z_{k-1}' &= z_k \quad \forall k=1,\ldots,n-1 \nonumber \\
    z_{n-1}' &= f(t,z_0,\ldots,z_{n-1}) \nonumber \\
    \Leftrightarrow z' &= 
        \begin{bmatrix}
          z_1 \\
          \vdots \\
          z_{n-1}\\
          f(t,z)
        \end{bmatrix} \eqqcolon f_{\text{ext}}(t,z). \label{eq: symmetry condition extended first order system}
\end{align}

The symmetry condition for a higher order ODE is, analogous to the symmetry condition for a first order ODE,
\begin{align}
    X^{(n)} &\left(y^{(n)} - f(t,y,\ldots,y^{(n-1)})\right) = 0, \label{eq: symmetry condition higher order system}
\end{align}
In the following, it is shown that the symmetry condition for the first order system in Equation~(\ref{eq: symmetry condition extended first order system}) is equivalent to the symmetry condition for the corresponding higher order system in Equation~(\ref{eq: symmetry condition extended first order system}). For the ease of notation $y$ is assumed to be scalar valued, but the proof can be easily extended to $y$ being vector valued.\\
Assume that $W \coloneqq \xi \partial_t + \eta_0 \partial_{z_0} + \ldots + \eta_{n-1}\partial_{z_{n-1}}$ is a generator of a Lie symmetry of ODE~(\ref{eq: symmetry condition extended first order system}). The corresponding symmetry condition reads
\begin{align}
    &W^{(1)}\left( z_k' - f_{\text{ext},k}(t,z) \right) = 0. \nonumber \\
    \Leftrightarrow &\frac{\mathrm{d}\eta_k}{\mathrm{d}t} - z_k'\frac{\mathrm{d}\xi}{\mathrm{d}t} - \eta_{k+1} = 0  \quad \forall k=0,\ldots,n-2 \label{eq:sym cond ext k<n-1} \\
    & \frac{\mathrm{d}\eta_{n-1}}{\mathrm{d}t} - z_{n-1}'\frac{\mathrm{d}\xi}{\mathrm{d}t} - Wf = 0. \label{eq:sym cond ext k=n-1}
\end{align}
Now note, that under the variable transformation $z_k \rightarrow y^{(k)}$ Equation~(\ref{eq:sym cond ext k<n-1}) corresponds to the prolongation defined in Equation~(\ref{eq: prolongation higher order}), as $\eta_k \rightarrow \eta^{(k)}$, the $k$-th prolongation with respect to the generator $X = \xi \partial_t + \eta \partial_y$. Setting $\frac{\mathrm{d}\eta_{n-1}}{\mathrm{d}t} - z_{n-1}'\frac{\mathrm{d}\xi}{\mathrm{d}t} \eqqcolon \eta_n$ in Equation~(\ref{eq:sym cond ext k=n-1}) and $z_n \coloneqq y^{(n)}$, Equation~(\ref{eq:sym cond ext k=n-1}) can be expressed as
\begin{align}
    \left( W + \eta_n \partial_{z_n} \right) \left( z_n - f \right) = 0.
\end{align}
This can be seen to be equivalent to Equation~(\ref{eq: symmetry condition higher order system}), as 
\begin{equation*}
    W + \eta_n \partial_{z_n}  = \xi \partial_t + \eta_0 \partial_{z_0} + \ldots + \eta_{n}\partial_{z_n} \rightarrow \xi \partial_t + \eta \partial_{y} + \ldots + \eta^{(n)}\partial_{y^{(n)}} = X^{(n)}.
\end{equation*}

\section{ODE Test Problems}
In order to demonstrate the benefit of our method, we compared ourselves with existing methods on ten example systems of ODEs. Table~\ref{tab:example_systems} displays theses systems along with a known generator.
All examples are systems of two ODEs, hence they cannot be tackled by Maple. 
Our method involves solving the system numerically. For this procedure, we used the Runge-Kutta algorithm on the start ranges and time intervals displayed in Table~\ref{tab:example_systems_range}.
Finally, we report the generators found by our method in Table~\ref{tab:example_systems_result}. We additionally checked, whether the generator condition was fullfilled numerically and symbolically.

\begin{longtable}{p{1.5cm}cc}
     Name&ODE $f(t,y)$&Generator $\eta^\star(t, y)$\\
     \toprule
     ODE1&$\begin{bmatrix}
    y_{1}(t + \frac{y_{2}}{y_{1}})^2\\
    t^2y_{1}
    \end{bmatrix}$&$\begin{bmatrix}
        y_1\\
        y_2
    \end{bmatrix}$\\\\
    ODE2&$\begin{bmatrix}
    y_{1}^2y_{2}\exp(y_{1}^{-1})\\
    t\exp(-y_{1}^{-1})
    \end{bmatrix}$&$\begin{bmatrix}
        y_1^2\\
        y_2
    \end{bmatrix}$\\\\
    ODE3&$\begin{bmatrix}
    ty_{1}(y_{2} - \log(y_{1}))\\
    t + y_{2} - \log(y_{1})
    \end{bmatrix}$&$\begin{bmatrix}
        y_1\\
        1
    \end{bmatrix}$\\\\
    ODE4&$\begin{bmatrix}
    t^{-1}(2y_{1} + y_{2}\exp(-y_{1}t^{-2}))\\ 
    y_{2}
    \end{bmatrix}$&$\begin{bmatrix}
        t^2\\
        y_2
    \end{bmatrix}$\\\\
    ODE5&$\begin{bmatrix}
    y_{1}(t - \log(y_{1})\tan(t))\\
    -y_{2}\log(y_{1})\tan(t) + y_{2}
    \end{bmatrix}$&$\begin{bmatrix}
        y_{1}\cos(t)\\
        y_{2}\cos(t)
    \end{bmatrix}$\\\\
    ODE6&$\begin{bmatrix}
    y_{1}(ty_{2}y_{1}^{-1} + 2\log(y_{1})t^{-1})\\
    2y_{2}\log(y_{1})t^{-1}
    \end{bmatrix}$&$\begin{bmatrix}
        t^2y_{1}\\
        t^2y_{2}
    \end{bmatrix}$\\\\
    ODE7&$\begin{bmatrix}
    y_{2}\exp(-0.5y_{1}^2t^{-2}))y_{1}^{-1} + 0.5y_{1}t^{-1}\\
    -0.5y_{1}^2y_{2}t^{-3}
    \end{bmatrix}$&$\begin{bmatrix}
        ty_{1}^{-1}\\
        y_{2}t^{-1}
    \end{bmatrix}$\\\\
    ODE8&$\begin{bmatrix}
    \exp(-t)\sin(y_{2})\\
    \exp(-t)\sin(y_{1})
    \end{bmatrix}$&$\begin{bmatrix}
        \sin(y_{2})\\
        \sin(y_{1})
    \end{bmatrix}$\\\\
    ODE9&$\begin{bmatrix}
    ty_{2}\sin(y_{1})\\
    t\sin(y_{1})
    \end{bmatrix}$&$\begin{bmatrix}
    y_{2}\sin(y_{1})\\
    \sin(y_{1})
    \end{bmatrix}$\\\\
    ODE10&$\begin{bmatrix}
    t\log(y_{2})\\
    ty_{1}^2
    \end{bmatrix}$&$\begin{bmatrix}
        \log(y_{2})\\
        y_{1}^2
    \end{bmatrix}$\\
\label{tab:example_systems}\\
\caption{Example systems used in this paper.}
\end{longtable}

\begin{longtable}{lcc}
     Name&Start Range&Time interval\\
     \toprule
     ODE1&$[1, 2]$&$[1, 2]$\\
     ODE2&$[1, 2]$&$[0, 1]$\\
     ODE3&$[1, 2]$&$[1, 2]$\\
     ODE4&$[1, 2]$&$[1, 2]$\\
     ODE5&$[1, 2]$&$[1, 2]$\\
     ODE6&$[1, 2]$&$[1, 2]$\\
     ODE7&$[1, 2]$&$[1, 2]$\\
     ODE8&$[-1, 0]$&$[-1, 0]$\\
     ODE9&$[-1, 0]$&$[-1, 0]$\\
     ODE10&$[1, 2]$&$[1, 2]$\\
     \caption{Settings used for numerically solving example ODEs. We used three time series per system to guarantee identifiability (see subsection~\ref{sec:identifiability}).}\\
    \label{tab:example_systems_range}\\
\end{longtable}

\begin{longtable}{cccc}
Problem&Generator&Numeric Loss&Zero Symbolic Loss\\
\toprule
ODE1&$\begin{bmatrix}2 y_{1}\\2 y_{2}\end{bmatrix}$&$1.41e-13$&$\begin{bmatrix}\text{True}\\\text{True}\end{bmatrix}$\\
\midrule
ODE2&$\begin{bmatrix}y_{1}^{2}\\y_{2}\end{bmatrix}$&$1.03e-17$&$\begin{bmatrix}\text{True}\\\text{True}\end{bmatrix}$\\
\midrule
ODE3&$\begin{bmatrix}2 y_{1}\\2\end{bmatrix}$&$1.87e-17$&$\begin{bmatrix}\text{True}\\\text{True}\end{bmatrix}$\\
\midrule
ODE4&$\begin{bmatrix}t^{2}\\y_{2}\end{bmatrix}$&$1.06e-20$&$\begin{bmatrix}\text{True}\\\text{True}\end{bmatrix}$\\
\midrule
ODE5&$\begin{bmatrix}0\\2 y_{2}\end{bmatrix}$&$5.02e-20$&$\begin{bmatrix}\text{True}\\\text{True}\end{bmatrix}$\\
\midrule
ODE6&$\begin{bmatrix}t^{2} y_{1}\\t^{2} y_{2}\end{bmatrix}$&$1.18e-15$&$\begin{bmatrix}\text{True}\\\text{True}\end{bmatrix}$\\
\midrule
ODE7&$\begin{bmatrix}\frac{t}{y_{1}}\\\frac{y_{2}}{t}\end{bmatrix}$&$1.09e-22$&$\begin{bmatrix}\text{True}\\\text{True}\end{bmatrix}$\\
\midrule
ODE8&$\begin{bmatrix}\sin{\left(y_{2} \right)}\\\sin{\left(y_{1} \right)}\end{bmatrix}$&$2.31e-22$&$\begin{bmatrix}\text{True}\\\text{True}\end{bmatrix}$\\
\midrule
ODE9&$\begin{bmatrix}y_{2} \sin{\left(y_{1} \right)}\\\sin{\left(y_{1} \right)}\end{bmatrix}$&$1.64e-24$&$\begin{bmatrix}\text{True}\\\text{True}\end{bmatrix}$\\
\midrule
ODE10&$\begin{bmatrix}\log{\left(y_{2} \right)}\\y_{1}^{2}\end{bmatrix}$&$8.46e-21$&$\begin{bmatrix}\text{True}\\\text{True}\end{bmatrix}$\\
\midrule
\label{tab:example_systems_result}\\
\caption{Results obtained by our method on the example systems.}
\end{longtable}

\subsection{Identifiability}
\label{sec:identifiability}
For a system of $n$ ODEs, the general solution has $n$ parameters~\citep{coddington56}[Theorem 4.4], which can be grouped to a vector $a$. If we want to find the particular solution for fixed initial conditions $(t_0,y_1)$, we look for the corresponding parameter values for the general solution. But this means every time series corresponds to one instantiation $a_k$ of parameters. Thus, if we only observe this single time series, we would not be able to identify that there are $n$ parameters but would assume them constant. With a second time series, where the initial conditions are chosen such that one parameter has a different value, we could identify that there is a parameter that can assume different values. Continuing this reasoning, we would need $n+1$ time series with parameter vectors $\{a_k\}_{k=1}^{n+1}$ such that there is no $n-1$-dimensional affine subspace that contains all parameter vectors. This independence condition on the parameters follows from the fact that the parameter space needs to be $n$-dimensional, regardless of the choice of parametrization. Hence, we need at least $n+1$ different time series to identify the symbolic form of an $n$-dimensional ODE.\\
This can be illustrated on the example of the following simple ODE
\begin{equation*}
\label{eq:non_ident_general}
     y'=\frac{2y}{t},
\end{equation*}
which has the general solution
\begin{equation*}
    y(t) = at^2
\end{equation*}
for some parameter $a$. A single time series $\{(t_i,y_i)\}_i$ corresponds to a choice of an initial condition $y(t_0)=c_0$, which is the same as fixing a constant $a(t_0,c_0) = c_0/t_0^2$. The specific time series, however, could have also originated from the ODE 
\begin{equation*}
\label{eq:non_ident_particular}
    y'=2a(t_0,c_0)t.
\end{equation*}
The two ODEs have a different set of solutions and different symmetries, a scaling symmetry $\eta^\star = y$ for the first ODE and a translation symmetry $\eta^\star = 1$ for the second ODE. Hence, using only one time series and finding the second ODE would lead to wrong predictions for other initial conditions. This is why we would need at least two time series with different parameter values $a$.\\
In conclusion, when we know the general solution of an ODE, the parameters can be chosen such that the ODE can, in theory, be identified from the time series. In practice, however, this is usually not possible as the above independence condition on the parameters cannot be checked from the initial conditions alone without knowing the ODE. One can argue that this is not an algorithmic problem but rather a problem for the user, who provides the time series as input. An algorithm could, in theory, return an ODE model which describes the input data, even though it cannot be guaranteed that this model does generalize to different initial conditions.
Hence, for all experiments we selected enough time series whenever we could control it to ensure identifiability of the true ODE.

\end{document}